\documentclass[10pt]{article}   	
\usepackage{array}
\usepackage{color}
\usepackage{fullpage}
\usepackage{graphicx}				
\usepackage{tikz}
\usepackage{tikz-qtree}
\usepackage{amssymb}
\usepackage[authoryear, round]{natbib}
\usepackage{lineno}
\usepackage{url}
\usepackage{verbatim}

\newcommand{\specialcell}[2][c]{%
 \begin{tabular}[#1]{@{}l@{}}#2\end{tabular}}

\date{}							

\title{Formal models of Structure Building in Music, Language and Animal Song}
\author{
\normalsize Willem Zuidema\\
\small ILLC, University of Amsterdam\\
\small P.O.Box 94242, 1090 CE  Amsterdam, Netherlands\\[.5\baselineskip]
\normalsize Dieuwke Hupkes\\
\small ILLC, University of Amsterdam\\
\small P.O.Box 94242, 1090 CE  Amsterdam, Netherlands\\[.5\baselineskip]
\normalsize Geraint Wiggins\\
\small School of Electronic Engineering and Computer Science\\
\small Queen Mary University of London, Mile End Road, London E1 4NS.U.K.\\[.5\baselineskip]
\normalsize Constance Scharff\\
\small Animal Behaviour, Freie Universit\"at Berlin, Takustra{\ss}e 6, 14195 Berlin, Germany\\
\normalsize Martin Rohrmeier\\
\small Institut f\"ur Kunst- und Musikwissenschaft, Technische Universit\"at Dresden, \\
\small August-Bebel-Stra\ss e 20, 01219 Dresden, Germany\\[.5\baselineskip]
}

\begin{document}

\maketitle

\begin{abstract}
Human language, music and a variety of animal vocalisations constitute ways of sonic communication that exhibit remarkable structural complexity.
While the complexities of language and possible parallels in animal communication have been discussed intensively, reflections on the complexity of music and animal song, and their comparisons are underrepresented.
In some ways, music and animal songs are more comparable to each other than to language, as propositional semantics cannot be used as as indicator of communicative success or well-formedness, and notions of grammaticality are less easily defined.
This review brings together accounts of the principles of structure building in language, music and animal song, relating them to the corresponding models in formal language theory, with a special focus on evaluating the benefits of using the Chomsky hierarchy (CH).
We further discuss common misunderstandings and shortcomings concerning the CH, as well as extensions or augmentations of it that address some of these issues, and suggest ways to move beyond.
\end{abstract}

\section{Introduction}

Human language, music and the complex vocal sequences of animal songs constitute ways of sonic communication that evolved a remarkable degree of structural complexity, for which - extensive research notwithstanding - completely satisfactory explanatory and descriptive models have yet to be found.
Formal models of structure have been most commonly proposed in the field of natural language, often building on the foundational work of Shannon and Chomsky in the 1940s and 1950s \citep{shannon1948, chomsky1956three}. Research in mathematical and computational linguistics has resulted in extensive knowledge of the formal properties of such models, as well as of their fit to phenomena in natural languages.
Such formal methods have been much less prevalent in modelling music and animal song.
Research using formal models of sequential structure has often focused on \textit{comparing} the structure of human language to that of learned animal songs, focusing particularly on songbirds, but also whales and bats \citep[e.g.,][]{doupe1999birdsong, bolhuis2010twitter, hurford2011origins, knoernschild2014male}. 
Such comparisons addressed aspects of phonology \citep[e.g.,][]{yip2006search, spierings2014zebra} and syntax \citep[e.g.,][]{berwick2011songs, ten2013analyzing, markowitz2013long, sasahara2012structural}, aiming to identify both species-specific principles of structure building and cross-species principles underlying the sequential organisation of complex communication sounds.
In such comparisons there has been a big focus on the role of `recursion' as a core mechanism of the language faculty in the narrow sense \citep{hauser2002faculty} and its uniqueness to both humans and human language. 

However, although recursion and the potential uniqueness of other features of language are important topics, it is certainly not the only relevant topic for comparative studies \citep[see also][]{fitch2005evolution,fitch2006biology, rothenberg2014investigation}.
Structurally and functionally, music, language and animal song not only share certain aspects but also have important differences.
A three-way comparison between language, music and animal songs and the techniques that are used to model and explain them, has the potential to benefit research in all three domains, by highlighting shared and unique mechanisms as well as hidden assumptions in current research paradigms.

In this chapter we present an overview of research considering structure building and sequence generation in language, music and animal song.
Our starting point is the work of Shannon and Chomsky from the 1940s and 1950s, which has been prominent in establishing a tradition of research in formal models of structure in natural language.
We discuss issues concerning building blocks, Shannon's \textit{n}-gram models and the Chomsky hierarchy (CH), as well as the limitations of both frameworks in relation to empirical observations from the biological and cognitive sciences.
We then proceed with discussing ways of addressing these limitations, including extending the CH with more fine-grained classes, the addition of probabilities and meaning representations to symbolic grammars, and replacing abstract symbols with numerical vectors.
At the end of the chapter, we reflect on what type of conclusions can be drawn from comparing and using these models and what impact this may have for future research.

\section{Building blocks and sequential structure}

Models for sequence generation highly depend on the choice of atomic units of the sequence.
Before considering models of structure building, we may first want to identify what the elementary building blocks are that sequences - be it in language, music or animal vocalisations - are built up from. 
This, however, turns out to be much more complicated than we might naively expect.

\subsection{Elementary units of models of language}
One of the classical universal `design features' of human language is duality of patterning \citep{hockett60sciam}, which refers to the fact that all languages show evidence of at least two combinatorial systems: one where meaningless units of sounds are combined into words and morphemes, and one where those meaningful morphemes and words are further combined into words, phrases, sentences and discourse.  
Although the two systems are not independent and arguably should not be considered this way, in this chapter we pragmatically focus mainly on the second combinatorial system that combines already meaningful units into larger pieces, because this is the target of the most heavily studied models of structure building in natural language.
Later in the chapter (in Section \ref{sec:discussion}) we briefly consider the interplay between the two systems.

But even when restricting ourselves to meaning-carrying units, it turns out to be far from trivial to identify phoneme, morpheme, syllable or word boundaries based on cues in the observable signal (i.e.\ a spectogram) alone \citep{liberman1967perception}.
The choice of elementary units of models for structure in language is therefore usually not based on features of the acoustic information but, rather, on semantic information accessible through introspection.
Most commonly, models considering structure in language are defined over \textit{words}.

\subsection{Building blocks of animal song}

Like language, animal songs combine units of sound into larger units in a hierarchical way, but the comparability of the building blocks and the nature of the hierarchical structure in language, music and animal song is not at all straightforward.
In particular, there are no clear analogues for words, phrases or even sentences in animal song \citep{bloom2004can, scharff2011evo}, and regardless of the approach taken to establish the smallest unit of the sequence (in bird song commonly referred to with the term `note' or `element') making decisions that are somewhat arbitrary seems unavoidable. 
A common way of identifying units in animal songs is to study their spectrogram and delineate units based on acoustic properties such as silent gaps \citep[e.g.,][]{isaac1963ordering, marler1984species, adam2013new, feher2009novo} or changes in the acoustic signal \cite[e.g.,][]{clark2008anna, payne1971songs}.
Additionally, evidence about perception and production of different acoustic structures is often used to motivate a particular choice of building blocks \citep[e.g.,][]{tierney2011motor, cynx1990experimental, franz2002respiratory, amador2013}.
Strikingly, choices regarding building blocks might also be made by studying patterns of recombination and co-occurrence \citep{podos1992organization, ten2013analyzing}, an observation that illustrates the interdependence of the choice of building blocks and models for structure building, an issue that we revisit in Section \ref{sec:discussion}.
For a detailed review of the different methods used to identify units in animal vocalisations, we refer to \citep{kershenbaum2014animal}.

\begin{figure}[!h]
\centering
\includegraphics{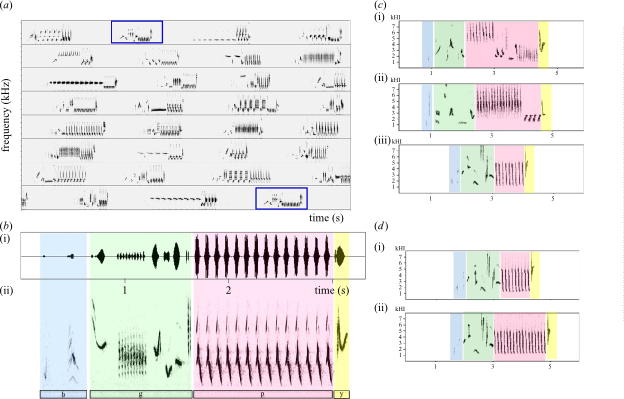}

\caption{Hierarchical organisation of nightingale song. Panel A depicts a spectrogram of ca 2 minutes of continuous nocturnal singing of a male nightingale.  Shown are 30 sequentially delivered unique songs. The 31st song is the same song type as the second one (blue frames).  On average a male has a repertoire of  180 distinct  song types, which can be delivered in variable but non-random order for hours continuously.  Panel B illustrates the structural components of one song type.  Individual sound elements are sung at different loudness (amplitude envelope on top) and are acoustically distinct in the frequency range, modulation, emphasis and temporal characteristics (spectrogram below).  Panel C illustrates the structural similarities in three different song types (a, b, c): Song types begin usually with one or more very softly sung elements (blue), followed by a sequence of distinct individual elements of variable loudness (green). All song types contain one or more sequences of loud note repetitions (red) and are usually ended by a single, acoustically distinct element (yellow). Panel D illustrates that the same song type can vary in the number of element repetitions in the repeated section (red). For details, see \citep{todt1998how}. Spectrograms modified from Henrike Hultsch (unpublished).}\label{fig:bird_song} 

\end{figure}

\subsection{Basic elements in music}

In music - aside from the lack of a (compositional) semantic interpretation - the complexity of the musical surface (i.e.\ an interplay of different features like rhythm, metric, melody and harmony\footnote{Although many of these aspects also occur in speech (cf. prosody), the structural aspect of human language appears to be easier to isolate.}), leaves an even larger spectrum of possible choices for building blocks.
Models can be defined not only over notes or chords, but also intervals and durations of notes, or other more complex features could be used as elementary units of the sequence.
Traditionally, much of the discussion of structure in music has focused on Western classical music and has built on building blocks of melody, voice-leading \citep[e.g.,][]{tymoczko2006geometry, callender2008generalized, quinn2011voice}, outer voices \cite[e.g.,][]{aldwell2010harmony}, harmony \citep{winograd1968linguistics, rohrmeier2007generative, rohrmeier2011towards}, combinations of harmony and voice-leading \citep{lerdahl1983, aldwell2010harmony, rohrmeier2015towards}, or complex feature combinations derived from monophonic melody \citep{conklin1995multiple, pearce2005} and harmony \citep{whorley2013multiple, rohrmeier2012comparing}.  

The choice of building blocks is thus a difficult issue in all three domains we consider, and any choice will have important consequences for the models of structure that can be defined over these building blocks. The fact that choices regarding the `units of comparison' may strongly affect the conclusions that can be drawn is frequently overlooked in the literature comparing birdsong, music and language.Nevertheless, it is often best to make pragmatic decisions about the building blocks in order to move on; as it turns out, some of the questions about building blocks can be addressed only after having considered models of structure (at which point applying \textit{model selection}, a topic we revisit later, can help to revisit choices about the building blocks).

\section{Shannon's \textit{n}-grams}

In the slipstream of his major work on information theory, \cite{shannon1948} introduced \textit{n}-gram models as a simple model of sequential structure in language. 
\textit{n}-grams define the probability of generating the next symbol in a sequence in terms of the previous (\textit{n-1}) symbols generated. 
When \textit{n=2}, the probability of generating the next word depends only on what the current word is, and the \textit{n}-gram model - called a `bigram' model in this case - simply models transition probabilities.
\textit{n}-gram models are equivalent to (\textit{n-1})th-order Markov models over the same alphabet.

\subsection{Probability estimation}

\textit{n}-gram probabilities can be estimated from a corpus using maximum likelihood estimation (or relative frequency estimation) \citep{jurafsky2000}.
In theory, the bigger the value of \textit{n}, the better one can predict the next word in a sentence, but in practice no natural language corpus is large enough to estimate the probabilities of events in the long zipfian tail of the probability distribution with relative frequency estimation.\footnote{This is an even bigger problem when trying to model bird song, where datasets are generally small and in many cases the number of possible transition probabilities - despite the comparably small number of elementary units - vastly exceeds the number of examples in the entire set of empirical data.} 
When human language is modelled, this problem is usually addressed by decreasing the probability of the counted \textit{n}-grams and reassigning the resulting probability mass to unseen events, a process called smoothing or discounting \citep{manning1999foundations}.
Smoothed \textit{n}-gram models have long been the state of the art for assigning probabilities to natural language sentences, and tlhough better performing language models \cite[in terms of modelling the likelihood of corpora of sentences, e.g.,][]{schwenk2005training, mikolov2012statistical} have been developed now, \textit{n}-gram models are still heavily used in many engineering applications in speech recognition and machine translation, due to their convenience and efficiency .

\subsection{\textit{n}-gram models of birdsong}

\textit{n}-gram models (often simple bigrams) have also been frequently applied to bird song \citep{isaac1963ordering, chatfield1970analyzing, slater1983, okanoya2004, briefer2010, markowitz2013long, samotskaya2016syntax} and music \citep{ames1989markov, pearce2012auditory}. 
For many bird species, bigrams in fact seem to give a very adequate description of the sequential structure. 
\cite{chatfield1970analyzing} studied the song of the cardinal and reported that a 3-gram (trigram) model modeled song data only marginally better than a bigram model, measured by the likelihood of the data under each of these models.
There is a single, small data set used for extracting \textit{n}-grams and measuring likelihood, which makes drawing furm conclusions from this classic analysis difficult.
More recent work with birds that were exposed to artificially constructed songs as they were raisedsuggests that transitional probabilities between adjacent elements are the most important factor in the organisation of the songs also in zebra finches and Bengalese finches \citep{lipkind2013stepwise}, although many other examples of bird song also require icher models \citep{ten2013analyzing,  ten2012revisiting, katahira2011, katahira2013}.

\subsection{\textit{n}-gram models for music}

In music, numerous variants of \textit{n}-gram models have been used, to model musical expectancy \citep{narmour1990, schellenberg1997simplifying, schellenberg1996expectancy, krumhansl1995music, eerola2003}, but also to account for the perception of tonality and key \citep[which has been argued to be governed by pitch distributions that correspond to a unigram model][]{krumhansl2004cognition, krumhanslkessler1983} and to describe melody and harmony \citep{conklin1995multiple, ponsford1999statistical, reis1999, pearce2005, rohrmeier2006towards, rohrmeier2012comparing, whorley2013multiple}.
In particular, in the domain of harmony, Piston's table of common root progressions \citep{piston1948} and Rameau’s theory (of the \emph{basse fondamentale}) \citep{rameau1971} may be argued to have the structure of a first-order Markov model (a bigram model) of the root notes of chords \citep{hedges2011exploring, temperley2001cognition}.
In analogy with the findings in bird song research, several music modelling studies find trigrams optimal with respect to modelling melodic structure \citep{pearce2004improved} or harmonic structure \citep{rohrmeier2012comparing}, although here too, the size of the datasets used is too small to draw firm conclusions.

In choosing the optimal value of \textit{n}, some additional aspects that play a role are usually not considered in language and animal song.
For instance, because of the interaction of melody with metrical structure, not all surface symbols have the same salience when forming a sequence, which could be an argument to - in the face of data sparsity - prefer a \textit{4}-gram model over a \textit{3}-gram model to model music with a three-beat metrical structure, as a 3-gram necessarily cannot capture the fact that the first beat of a bar is, in harmonic terms, more musically salient than the other two \citep{ponsford1999statistical}.
More generally, the interaction between different single-stream features in music forms a challenge for \textit{n}-gram models, an aspect that is not as inescapable when modelling language and animal song \citep[but see][for an interesting, multi-stream pattern in zebra finch vocalisations and dance]{ullrich2016}. 
One model that addresses this problem by combining \textit{n}-gram models over different features and combined feature-spaces was proposed in \cite{conklin1995multiple}. 

\section{The classical Chomsky hierarchy}
\label{sec:chomskyhierarchy}

Shannon's \textit{n}-grams are simple and useful descriptions of some aspects of local sequential structure in animal communication, music and language. 
It is however often argued that they are unable to model certain key structural aspects of natural language.
In theoretical linguistics, \textit{n}-grams, no matter how large their \textit{n}, were famously dismissed as useful models of syntactic structure in natural language in the foundational work of Noam Chomsky from the 1950s \citep{chomsky1956three}.
In his work, Chomsky argued against incorporating probabilities into language models; in his view, the core issues for linguists concern the symbolic, syntactic structure of language.
He proposed an idealisation of natural language where a language is conceived of as a potentially infinite set of sentences, and a sentence is simply a sequence of words (or morphemes). By systematically analyzing the ways in which such sets of sequences of words could be generated, Chomsky discovered a hierarchy of increasingly powerful grammars, relevant for both linguistics and computer science, that has later been named the `Chomsky Hierarchy' (CH).

\subsection{Four classes of grammars and languages}

In its classical formulation, the CH distinguishes four classes of grammars and their corresponding languages: regular languages, context-free languages, context-sensitive languages, and recursively enumerable languages. 
Each class contains an infinite number of sets, and is strictly contained in all classes that are higher up in the hierarchy: every regular language is also context-free, every context-free language is also context-sensitive, and every context-sensitive language is recursively enumerable.
When probabilities are stripped off, (\textit{n}--grams correspond to a proper subset of the regular languages.

\subsection{The Chomsky hierarchy and cognitive science}

For cognitive science, the relevance of the hierarchy comes from the fact that the four classes can be defined by the kinds of rules that generate structures as well as by the kind of computations needed to parse the sets of sequences in the class (the corresponding formal automaton).
Informally, regular languages are the sets of sequences that can be characterised by a ``flowchart'' description, which corresponds to a finite-state automaton or FSA. 
Regular languages can be straightforwardly processed (and generated) from left to right in an incremental fashion.
Crucially, when generating or parsing the next word in a sentence of a regular language, we only needs to know where we currently are on the flowchart, not how we got there (for an example see Figure \ref{fig:FSA}).

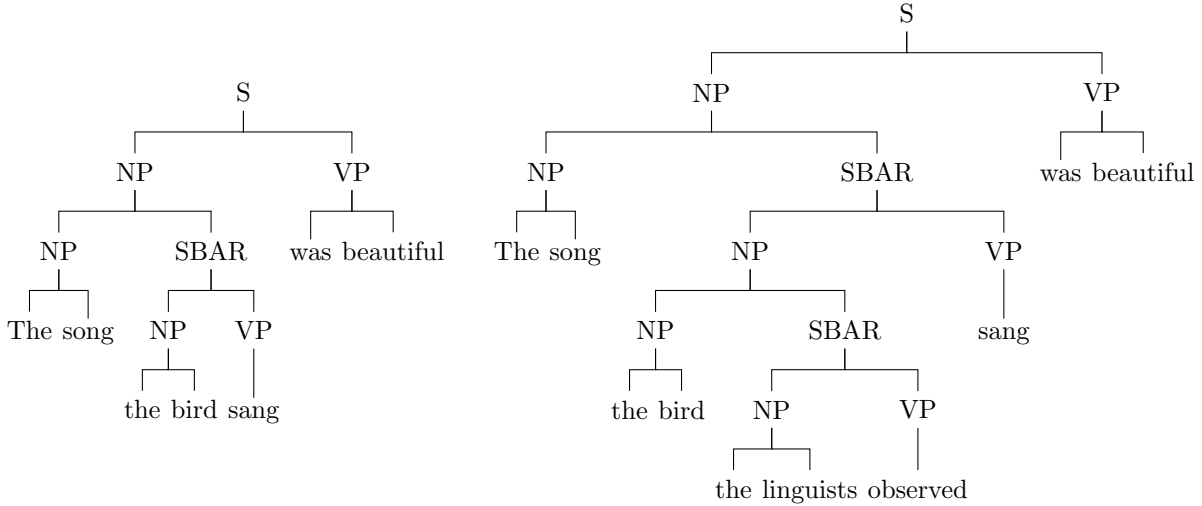
\begin{figure}[!h]

\begin{tabular}{m{11cm}cc}
The sentences \textit{The song the bird sang was beautiful} and \textit{The song the bird the linguists observed sang was beautiful} are examples of sentences with center embedding in English.
The latter sentence can be derived from the start symbol S by subsequently applying rules 1, 2a, 2b, 3, 2a, 2c, 2a, 2d, 4c, 4b, 4a.
(Note that traditionally, the analysis of the sentence contains a so called trace connecting the VP to its subject; left out here for clarity).
&& \specialcell{
(1) S $\rightarrow$ NP VP\\
(2a) NP $\rightarrow$ NP SBAR\\
(2b) NP $\rightarrow$ the song\\
(2c) NP $\rightarrow$ the bird\\
(2d) NP $\rightarrow$ the linguists\\
(3) SBAR $\rightarrow$ NP VP\\
(4a) VP $\rightarrow$ was beautiful\\
(4b) VP $\rightarrow$ sang\\
(4c) VP $\rightarrow$ observed\\
}
 
\end{tabular}

\vspace{3mm}

\begin{tabular}{cc}

    \specialcell{
    \begin{tikzpicture}[scale=1.0]
    \tikzset{level 1+/.style={sibling distance=-1mm}}
    \tikzset{edge from parent/.style= {draw, edge from parent path={(\tikzparentnode.south) -- +(0,-8pt) -| (\tikzchildnode)}}}

    \Tree [.S [.NP [.NP The song ] [.SBAR [.NP the bird ] [.VP sang ] ] ] [.VP was beautiful ] ] 
\end{tikzpicture}
}
&

\specialcell{
\begin{tikzpicture}[scale=1.0]
    \tikzset{level 1+/.style={sibling distance=-1mm}}
    \tikzset{edge from parent/.style= {draw, edge from parent path={(\tikzparentnode.south) -- +(0,-8pt) -| (\tikzchildnode)}}}

    \Tree [.S [.NP [.NP The song ] [.SBAR [.NP [.NP the bird ] [.SBAR [.NP the linguists ] [.VP observed ] ] ] [.VP sang ] ] ] [.VP was beautiful ] ]

\end{tikzpicture}
}

\end{tabular}

\caption{Center embedding in English.}\label{fig:center_embedding}

\end{figure}

At all higher levels of the CH, some sort of memory is needed by the corresponding formal automaton that recognises or generates the language. 
The next level up in the classical CH are context-free languages (CFL's), generated by context-free grammars (CFG's), equivalent to so-called push-down automata, that employ a simple memory in the form of a stack.
CFG's consist of (context-free) rewrite rules that specify which symbols (representing a category of words or other building blocks, or categories of phrases) can be rewritten to which list of symbols. 
Chomsky observed that natural language syntax allows for nesting of clauses (center embedding), and argued that finite-state automata are incapable of accounting for such phenomena. 
In contrast, context-free grammars can express multiple forms of nesting as well as forms of counting elements in a sequence. 
An example of such nesting, and a context-free grammar that can  describe it, is given in Figure \ref{fig:center_embedding}.

\subsection{Using the Chomsky hierarchy to model music}

The success of the CH in linguistics and computer science and Chomsky's demonstration that natural language syntax is beyond the power of finite-state automata has influenced many researchers to examine the formal structures underlying animal song and music (though there is no comprehensive comparison of models in either domain in terms of the CH yet). 
In music, there appears to be evidence for a number of nontrivial structure building operations at work that invite an analysis in terms of the CH or related frameworks. While more cross-cultural research is necessary, key structural operations that we can already identify include repetition and variation\citep{margulis2014repeat}, element-to-element implication (e.g.\ note-note, chord-chord) \citep{narmour1990, huron2006}, hierarchical organisation and tree structure, and nested dependencies and insertions \citep[e.g.,][]{widdess1981aspects, jackendoffLerdahl2006}). 
Most of these operations are more naturally expressed using CFGs than with FSAs, and indeed a rich tradition that emphasises hierarchical structure, categories and, particularly, recursive insertion and embedding exists to characterise Western tonal music \citep{winograd1968linguistics, keiler1978bernstein, keiler1983some, lerdahl1983, narmour1990,  steedman1984generative, steedman1996blues,de2009modeling, rohrmeier2007generative, rohrmeier2011towards, wilding2014robust, rohrmeier2015towards}. 

\begin{figure}[h!]
    \centering
    \includegraphics{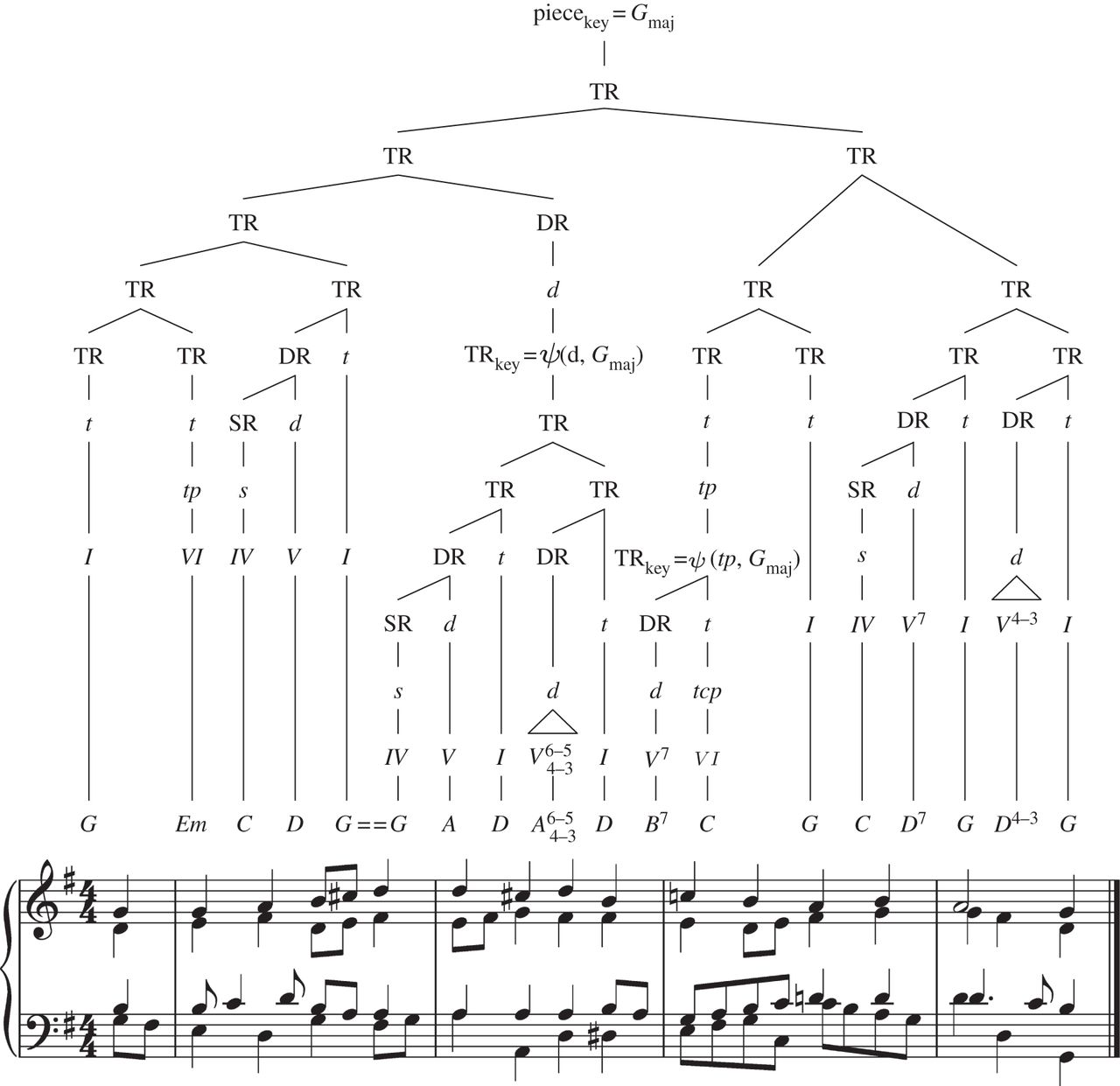}
    \caption{Analysis of Bach’s chorale "Ermuntre Dich, mein schwacher Geist" according to the GSM proposed by Rohrmeier\citep{rohrmeier2011towards}. The analysis illustrates hierarchical organisation of tonal harmony in terms of piece (\textit{piece}), functional regions (\textit{TR}, \textit{DR}, \textit{SR}), scale-degree (roman numerals) and surface representations (chord symbols). The analysis further exhibits an instance of recursive center-embedding in the context of modulation in tonal harmony. The transitions involving $TR_{key=\psi(x, ykey)}$ denote a change of key such that a new tonic region (TR) is instantiated from an overarching tonal context of the tonal function $x$ in the key $ykey$.} \label{fig:center_embedding_music} 
\end{figure}

However, music unlike language, does not convey propositional semantics. 
The function of the proposed hierarchical structures therefore cannot be the communication of a hierarchical, compositional semantics \citep{slevc2011meaning}, and one cannot appeal to semantics or binary grammaticality judgments to make the formal argument that language is trans-finite-state. 
Rather, a common thread in research about structure in music is that at any point in a musical sequence listeners are computing expectations about how the sequence will continue, regarding timing details and classes of pitches or other building blocks \citep{huron2006, rohrmeier2012predictive}. 
Composers can play with these expectations: meet expectations, violate them, or even put them on hold. In this play with expectations lie both the explanation for the existence of nested context-free structure in music and the way to make a more-or-less formal argument to place music on the CH. 
This is because the fact that an event may be prolonged \citep[i.e.\ extended through another event; an idea originating with][]{schenker1935} and events may be prepared or implied by other events, creates the possibility of having multiple and recursive preparations.
Employing an event as a new tonal center (musical modulation) could be formally interpreted as an instance of recursive context-free embedding of a new diatonic space into an overarching one (somewhat analogous to a relative clause in language) \citep{rohrmeier2007generative, rohrmeier2011towards, hofstadter1980}, which provides a motivation of the context-freeness of music through complex patterns of musical tension \citep{lerdahlKrumhansl2007X, lehneRohrmeierKoelsch2013}.
Figure \ref{fig:center_embedding_music} shows an example of a syntactic analysis of the harmonic structure of a Bach chorale that illustrates an instance of recursive center-embedding in the context of modulation. 

Although the fact that composers include higher-level structure in their pieces is uncontroversial, whether listeners are actually sensitive to such structures in day-to-day listening is a debated topic \citep{heffner2015prosodic, koelsch2013processing, farbood2015neural}. 
An alternative potential explanation for the existence of hierarchical structure in music could be found in the notation system and tradition of formal teaching and writing, a factor that may even be relevant for complexity differences in written and spoken languages in communities that may differ with respect to their formal education \cite{zengel1962literacy}.
However, there are also analytical findings that suggest that principles of hierarchical organisation may be found in classical North Indian music \citep{widdess1981aspects} that is based on a tradition of extensive oral teaching. 
More cross-cultural research on other cultures and structures in more informal and improvised music is required before conclusions may be drawn concerning structural complexity and cross-cultural comparisons. 

\subsection{The complexity of animal vocalisations}

In animal vocalisations, there is little evidence that non-human structures or communicative abilities (in either production or in reception) exceed finite-state complexity.
However, a number of studies have examined abilities to \textit{learn} trans-finite-state structures \citep[e.g.,][]{fitchHauser04science, lipkind2013stepwise, chen2015artificial, chen2016zebra}. 
Claims have been made - and refuted - that songbirds are able to learn such instances of context-free-structures (see \citep{gentner2006, abeWatanabe2011}; and respective responses \citep{vanHeiningen2009, ten2014phonetic, zuidema2013a, beckersBolhuisOkanoyaBerwick2012, corballis2007recursion}). 
Hence further targeted research with respect to trans-finite-stateness of animal song is required to shed light on this question.\footnote{No research studies have so far addressed the question whether the power of context-free grammars to express counting and numbered relationships between elements in musical or animal song sequences are required in real-world materials.}
By contrast, a number of studies argues for implicit acquisition of context-free structure (and even (mildly) context-sensitive structure in humans in abstract stimulus materials from language and music \citep{jiang2012, udden2012, rohrmeierFuDienes2012, rohrmeier2008statistical, li2013, rohrmeierRebuschat2012, kuhnDienes2005}. 

\section{Practical limitations of the Chomsky hierarchy}

It has turned out to be difficult to empirically decide where to place language, music and animal song on the CH, due to a number of different but related issues.
One of the more easily addressable problems concerns the finegrainedness of the levels of the CH.
It was observed in many studies that (plain) \textit{n}-grams are inadequate models of the structure of the vocalisations of several bird species on both the syllable and phrase \citep[e.g.,][]{markowitz2013long, jin2011compact, okanoya2004, katahira2011} and song \citep{todt1996acquisition, slater1983sequences} level.
Although \textit{n}-gram models seem to suffice for modelling songs of for instance mistle thrushes \citep{isaac1963ordering, chatfield1970analyzing} and zebra finches \citep{zann1996zebra}, richer models are needed to characterise the vocalisations of Bengalese finches \citep[e.g.,][]{katahira2011, katahira2013}, blackbirds \citep{todtWolffgram1975, ten2013analyzing} and other birds singing complex songs \citep[see][for a review]{kershenbaum2014animal, ten2012revisiting}.
However, this difference in complexity is not captured by the CH, as the complexer 
models proposed \citep[e.g., hidden Markov models,][]{rabiner1986introduction} are still finite-state models that fall into the lowest complexity class of the CH: regular languages. 

A similar issue occurs on higher levels of the hierarchy, when one tries to establish the formal complexity of natural languages such as English.
It was noticed already in the 1980's that some natural languages seem to display structures that are not adequately modeled by CFG's \citep{shieber1987, huybregts1984weak, culy1985complexity}.
However, the class of context-sensitive languages - one level up in the hierarchy - subsumes a much larger set of complex generalisations, many of which are never observed in natural language.

\subsection{Adding extra classes}

Both issues can be addressed by extending the CH with more classes.
Rogers and colleagues \citep{jaegerRogers2012} do so on the smallest level of the hierarchy, describing a hierarchy of sub-regular languages that contains the set of strictly local (SL) languages, which constitute the non-probabilistic counterpart of \textit{n}-gram models.
In the 1990's, \cite{joshi1990convergence} pointed out that a number of linguistic formalisms (e.g.\ tree-adjoining grammars \citep{joshi1975tree} or combinatorial categorial grammar\footnote{Which was also proposed in the music domain to describe harmony in jazz \citep{steedman1996blues}.}
\citep{steedman00book}) proposed to address the  apparent inadequacy of CFG's to model natural language, are formally equivalent with respect to the class of languages they are describing.
These languages, collectively referred to as mildly context-sensitive languages \citep[MCSL's,][]{joshi85mcsl}, can be roughly characterised by the fact that they are a proper superset of context-free languages, can be parsed in polynomial time, capture only certain kinds of dependencies and have constant growth property \citep{joshi1990convergence}.

\begin{figure}
\centering
    \includegraphics[scale=0.8]{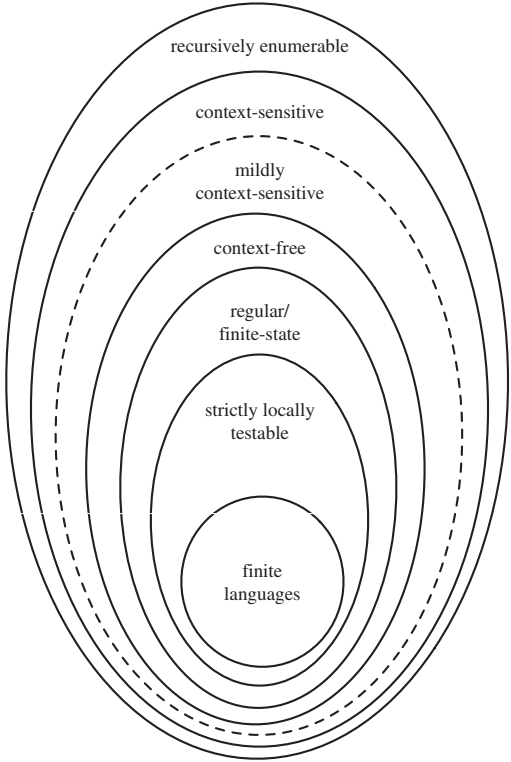}
    \caption{The extended Chomsky hierarchy.}\label{fig:extended_CH}
\end{figure}

\subsection{Empirically establishing the complexity of different languages}

Orthogonal to this granularity problem, is the more difficult problem of empirically evaluating membership of a class on the (extended) CH. 
The mere fact that a set of sequences can be built by a grammar from a certain class does not constitute a valid form of argument to place the system in question at the level on the (extended) CH. This is because a system lower in the hierarchy can approximate a system higher in the hierarchy with arbitrary precision by simply listing instances. 
For instance, Markov models are successfully used to describe some statistical features of corpora of music, \citep[e.g.,][]{tymoczko2003, deClercqTemperley2011, rohrmeier2008statistical, huron2006}, but crucially, this fact does not imply that a Markov model is also the best model.
The best model might be a context-free model that involves a single rule that captures a generalisation not captured by many specific nodes in the HMM.

To drive home this point further, consider again the example of center embedding described in Section \ref{sec:chomskyhierarchy}.
Note that arbitrarily deep center embeddings do not occur in practice \citep[and even center embeddings of very limited depth are rarely observed and are shown to be incomprehensible for most humans; see e.g.,][]{miller1964free, stolz1967study}. 
In any real world data, there is thus always only a finite (and, in fact, relative small) number of center embeddings; this finite set is easily modelled with an FSA that contains a different state for each depth. 
A finite-state account, however, loses a generalisation: different states lead to the same types of sequences, and we suddenly have a strict upperbound on the depth of possible embeddings.

\begin{figure}
    \begin{minipage}[c][7cm]{0.6\textwidth}
         \includegraphics{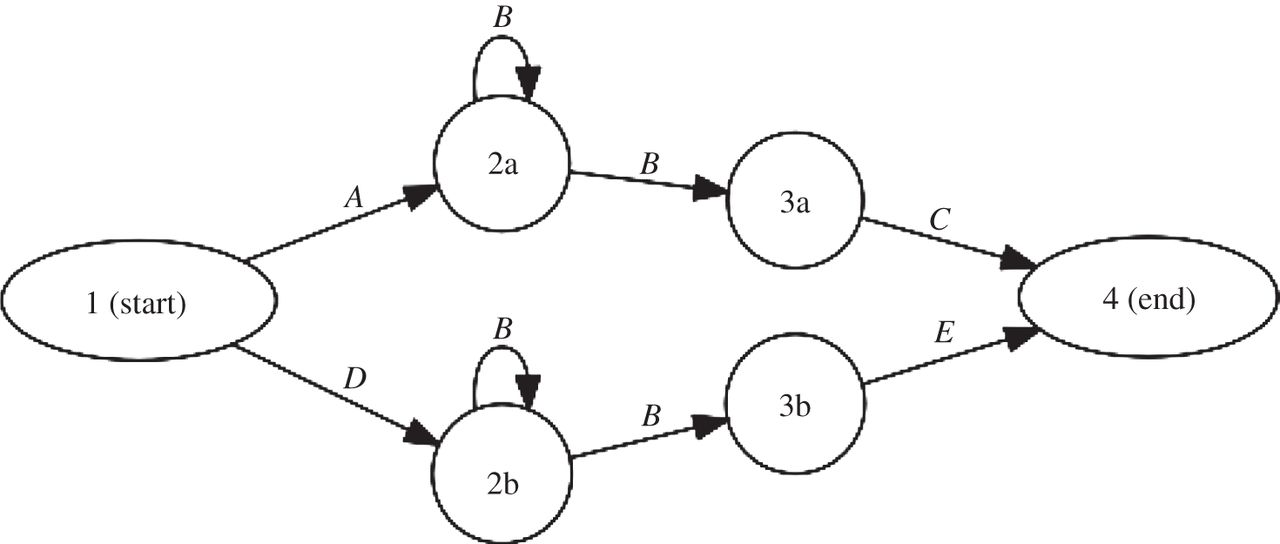}
     \end{minipage} \hfill
     \begin{minipage}[c][7cm]{0.3\textwidth}
         {\large\bfseries
        \begin{tabular}{l}
        \\
          S $\rightarrow$ AXC\\
          S $\rightarrow$ DXE\\
          X $\rightarrow$ BX\\
          X $\rightarrow$ B
      \end{tabular}}
    \end{minipage}
\caption{A finite-state automaton and a context-free grammar generating a repertoire consisting of two sequences: $AB^nC$ and $DB^nE$ (with \textit{n>0}). Note that the finite-state automaton is redundant in comparison with the CFG in the way that it contains multiple instances of the same structure $B^n$.}\label{fig:FSA}
\end{figure}

A similar issue arises when long-distance dependencies are used to prove the inadequacy of finite-state models.
For instance, when a bird sings songs of the structure $AB^nC$ and $DB^nE$, a long-distance dependency between \textit{A} and \textit{C} and between \textit{D} and \textit{E} can be observed, but the songs can be easily modeled with FSAs (see Figure \ref{fig:FSA}) by just assuming two different (hidden) states from which the B’s are generated: one for the condition starting with \textit{A} and ending with \textit{C}, and one for the other. 
This explains why some efforts to empirically demonstrate the context-freeness of bird song or music may not be convincing from a formal language theory perspective if they are based on just demonstrating a long-distance dependency. 
However, a long-distance dependency does have consequences for the underlying model that can be assumed in terms of its strong generative capacity (i.e. the set of \textit{structures} it can generate) and compressive power: in the example shown in Figure \ref{fig:FSA}, we were forced to duplicate the state responsible for generating \textit{B}, in fact we require \textit{2m} states (where \textit{m} is the number of non-local dependency pairs, such as $A\ldots C$ or $D\ldots E$, that need to be encoded). 
Therefore, if there are multiple (finite), potentially nested non-local dependencies, the number of required states grows exponentially, which is arguably unsatisfactory when considering strong generative capacity arguments \citep[see also the comparable argument regarding the implicit acquisition of such structures in][]{rohrmeierFuDienes2012, rohrmeier2014implicit}.
If the intervening material in a long-distance dependency is very variable, even if not technically unbounded, considerations of parsimony, strong-generative capacity, elegant structure-driven compression and considerations of efficiency provide strong reasons to prefer a model other than the minimally required class in the CH, or a different type of model altogether.

\subsection{Learnability}

A third type of problem concerns the learnability of certain types of structures from examples, and the complexity of the inference process that this requires.
A common paradigm to probe the type of generalisations made by humans and other species is to generate a sequence of sentences from an underlying grammar and study how well test subjects learn them in an artificial grammar learning (AGL) experiment \citep{reber1967implicit, pothos2007theories}. 
Such experiments show robustly, for instance, that humans are able to distinguish grammatical from ungrammatical sentences generated by a CFG without having very explicit knowledge of the `rules' they are using to make this distinction \citep{fitch2012pattern}. 
Many examples of AGL experiments with birds \citep[not even necessarily songbirds; see e.g.,][for an AGL study with pigeons]{herbranson2008artificial} and other animals \citep[such as rats or monkeys,][respectively]{murphy2008rule, wilson2013auditory} can be found in the literature. 

The results of AGL experiments remain difficult to interpret, as the inference procedures (and their complexity) to learn even relatively simple structures from examples are not well understood. 
Reviews of the ample number of AGL studies with both humans and non-human animals, as well as more formal accounts of their interpretability can be found in \citep{fitch2012pattern, ten2016assessing} and \citep{fitch2012artificial, pothos2007theories}, respectively.

\subsection{Relating the Chomsky hierarchy to cognitive and neural mechanisms}

Another class of arguments to move beyond the confines of the CH comes from considering its relation with cognition.
Historically, the CH is a theoretical construct that organises types of structures according to different forms of rewrite rules, and it has little immediate connection with cognitive motivations or constraints (such as limited memory)
Although it has been extensively used in cognitive debates on human and animal cognitive capacities, the CH may be quite fundamentally unsuitable for informing cognitive or structural models that capture frequent structures in language, music and animal song.
This may be a surprising claim, given the long and proud history of the Chomsky hierarchy, the fact that its classes are organised in mutual superset relations and the fact that the top level contains all recursively enumerable languages : everything has its place on the Chomsky hierarchy.
However, all the mathematical sophistication of the classes on the CH does not motivate their reification in terms of mental processes, cognitive constraints or neural correlates.
A metaphor might help drive this point home.
All squares on a chess-board can be reached by a knight in a finite number of steps.
We can therefore compute the distance between two fields in terms of the number of moves a knight needs.
This metric is universal in some sense (it applies to any two squares), but in general it is unhelpful, because its primitive operation (the knight's jump) is not representative for other chess pieces.
Similarly, the CH's metric of complexity is universal, but its usefulness is restricted by the primitive operations (rewrite operations) it assumes.

A well-known issue that further illustrates this point is the fact that repetition, repetition under a modification (such as musical transposition), and cross-serial dependencies constitute types of structures that require quite complex rewrite rules \citep[see also the example of context-sensitive rewrite rules expressing cross-serial dependencies in][]{rohrmeier2014implicit}, where such phenomena, in contrast, are frequent forms of form-building in music and animal song.
Mechanisms that can recognise and generate context-free languages are not limited to rewrite rules or even phrasal constituents \citep[consider e.g.\ dependency grammars][that describe the structure of a string in terms of binary connections between its elements]{tesniere1959elements}, and the mismatch between the simplicity of repetitive structures and the high CH class it is mapped onto might be one of many motivations to consider different types of models.

Other motivations to move beyond the confinements of the CH lie in the modelling of real-world structures that undermine some of the assumptions of the CH. 
Generally, the observation that music involves not only multiple parallel streams of voices, but also correlated streams of different features and complex timing, constitutes a theme that receives considerable attention in the domain of music cognition, but it does not easily match with the principles that underlie the CH, which is based on modelling a single sequence of words.
Similarly, one can argue that the CH is incapable of dealing with several essential features and characteristics of language, such as the fact that language is primarily used to convey messages with a complex semantic structure and the gradedness of syntactic acceptability \citep{aarts2004modelling, sorace2005gradience}.

In summary, the CH does not constitute an inescapable a priori point of reference for all kinds of models of structure building or processing, but it has inspired research in terms of a framework that allowed the comparison of different models and formal negative arguments against the plausibility of certain formal languages or corresponding computational mechanisms.
Such formal comparison and proofs should inspire future modelling endeavours, yet better forms of structural or cognitive models may involve distinctions orthogonal to the CH and may be designed and evaluated in the light of modelling data and its inherent structure as well as possible. 

\section{Moving towards different types of models}

Considering the challenges we have mentioned, what are some different aspects that new models of structure building and corresponding cognitive models should take into account?
In the slew of possible desiderata for new models, we observe two categories of requirements that such models should address.

\subsection{Modelling observed data}

The first category regards the suitability of models to deal with the complexity of actual real-world structures, which includes being able to deal with graded syntactic acceptability, but also handling semantics and form-meaning interactions. 
One main aspect that is particularly relevant is the notion of grammaticality or wellformedness.
The CH relies quite strongly on this notion for establishing and testing symbolic rules, but the idea that grammaticality is a strictly binary concept is problematic in the light of real-world data.
Even if the underlying system would prescribe so in theory, models should be able to account for the fact that in practice grammaticality is graded rather than binary \citep[e.g.,][]{abney1996}.\footnote{Related to this, due to to the fact that our knowledge of possible rules in language is largely implicit, it is not always easy to even to agree on the structural analysis of a sentence. Even when the formalism used for analysis is fixed, trained linguists are not always in agreement on which tree exactly should be assigned to a certain sentence, see e.g.\ \cite{berzak2016bias, brants2003syntactic}.}
In the case of music, it is not clear whether ungrammatical or irregular structures are clear-cut or distinguished in agreement by non-expert or expert subjects.
This problem is even more prominent in the domain of animal songs, where introspection cannot be used to assess the grammaticality of sequences or the salience of proposed structures, and research can typically be based only on so-called positive data, examples conforming with the proposed rules.
It is significantly more difficult to establish the validity and extension of rules in absence of negative data - i.e.\ where humans or animals explicitly reject a malformed sequence - which is hard to obtain in case of animal research. 

\subsection{Evaluation and comparison}

A second category of requirements concerns the evaluation and comparison of different models.
As we have pointed out, the CH is not particularly useful for selecting or even distinguishing models based on empirical data, as it provides no means to quantify the fit of a certain model with observed data.
To overcome this problem, new models should include some mechanism that allows the modeller to evaluate which model better describes experimental data, for instance by evaluating the agreement of their complexity judgments with empirical findings from the sentence processing literature \citep[e.g.,][]{engelmann2009processing, vasishth2010short, gibson1999memory}, their assessment of the likelihood of observed or made-up sequences, or by evaluating their predictive power.
The last method of evaluating models seems particularly suitable for music, where empirical data are often focused around the expectations listeners are computing about how the sequence will continue.
Further considerations to prefer one model over another could be grounded in descriptive parsimony or minimum description length \citep{mavromatis2009minimum}.

In the remainder of this chapter, we discuss three important extensions of the CH that address some of the previously mention issues.

\section{Dealing with noisy data: adding probabilities}

An important way to build better models of cognition and deal with issues from both above mentioned categories comes from reintroducing the probabilities that Chomsky abandoned along with his rejection of finite-state models.
A hierarchy of probabilistic grammars can be defined that is analogous to the classical (and extended) CH and exhibits the same expressive power.
We already mentioned that augmenting the automata generating SL languages yields \textit{n}-gram models, whereas the probabilistic counterpart of an FSA is a hidden Markov model (HMMs). 
Similarly, CFGs and CSGs can be straightforwardly extended to probabilistic CFGs (PCFGs) and probabilistic CSGs (PCSG's), respectively. 

Adding probabilities to the grammars defined in the CH addresses many of the issues mentioned above.
Probabilistic models can deal with syntactic gradience by comparing the likelihood of observing particular sentences, songs or musical structures (although accounting for human graded grammaticality judgments is not easy, see e.g., \citealp{lau2015unsupervised}).
Furthermore, they lend themselves well to information-theoretic methodologies such as model comparison, compression or minimum description length \citep{grunwald2007minimum, mackay2003information}. 
Probabilities allow us to quantify degrees of fit, and thus select models in a Bayesian model comparison paradigm by selecting the model with the posterior probability given the data and prior beliefs or requirements. 
In addition, probabilistic models permit defining a probability distribution over possible next words, notes or chords in a sequence, which matches well with many experimental data about sentence and music processing.

The use of probabilistic models is widespread in both music, language and animal song.
Aside from the previously mention \textit{n}-gram models, frequently applied in all three domains, more expressive probabilistic models have also been widely used.
Pearce's IDyOM model - an extension of the multiple feature \textit{n}-gram models proposed by \citeauthor{conklin1995multiple} - has been shown to  be successful in the domains of both music and language \citep{pearce2012auditory}
Recent modelling approaches generalised the notion of modelling parallel feature streams into dynamic Bayesian networks that combine the advantages of HMMs with modelling feature streams \citep{murphy2002dynamic, rohrmeier2012comparing, raczynski2013dynamic,paiement2008}.

In general, HMMs - which assume that the observed state is generated by a sequence of underlying (hidden) states that emit surface symbols according to a given probability distribution  \citep[for a comprehensive tutorial see][]{rabiner1989tutorial} - have been used extensively to model sequences in music \citep[e.g.,][]{rohrmeier2012comparing, mavromatis2005hidden, raphael2004functional} and animal song \citep[e.g.,][]{katahira2011,jin2011compact}. 

HMMs are also frequently practiced in modelling human language, although their application is usually limited to tasks regarding more shallow aspects of structure, such as part-of-speech tagging \citep[e.g.,][]{brants2000tnt} or speech recognition \citep[e.g.,][]{rabiner1993fundamentals, huang1990hidden}).
For modelling structural aspects of natural language, researchers usually resort to probabilistic models higher up the hierarchy, such as PCFG's \citep[e.g.,][]{petrov2007improved}, lexicalised tree-adjoining Grammars \citep{joshi1975tree} or Combinatory Categorial Grammars \citep{steedman00book}.

\section{Dealing with meaning: adding semantics}

One crucial aspect of human language of language that does not play a role in the CH is semantics. 
Chomsky's original work stressed the independence of syntax from semantics, but that does not mean that semantics is not important for claims about human uniqueness or structure building operations in language, even for linguists working within a `Chomskian' paradigm. 
\cite{berwick2011songs}, for instance, uses the point that bird song crucially lacks underlying semantic representations to argue against the usefulness of bird song as a comparable model system for human language. 
Their argument is that in natural language the trans-finite-state structure is not some idiosyncratic feature of the word streams we produce, but something that plays a key role in mediating between thought (the conceptual-intentional system in Chomsky's terms) and sound (the articulatory-perceptual system). 
Note that while the relevance of the interlinkedness of thought and sound in language is an important point, we are not sure on which evidence \cite{berwick2011songs}  ground their statement that birds song lacks semantic representations.  
\subsection{Transducers}

Crucially, the conceptual-intentional system is also a hierarchical, combinatorial system \citep[most often modeled using some variety of symbolic logic, most famously the system of ][]{montague1970universal}.
From that perspective, grammars from the (extended) CH describe only one half of the system; a full description of natural language would involve a transducer that maps meanings to forms and vice versa  \citep[e.g.,][]{jurafsky2000, zuidema2013b}. 
For instance, finite-state grammars can be turned into finite-state transducers, and context-free grammars into synchronous context-free grammars. 
All of the classes of grammars in the CH, have a corresponding class of transducers \citep[see][for an overview]{knight2005overview}. 
Depending on the type of interaction we allow between syntax and semantics, there might or might not be consequences for the set of grammatical sentences that a grammar allows if we extend the grammar with semantics. 
In any case, the extension is relevant for assessing the adequacy of the combined model - e.g.\ we can ask whether a particular grammar supports the required semantic analysis - as well as for determining the likelihood of sentences and alternative analyses of a sentence. 

\subsection{Semantics in music}

Whether we need transducers to model structure building in animal songs and music is a question that remains to be answered.
There have been debates about forms of musical meaning and its neurocognitive correlates. 
A large number of researchers in the field agree that music may feature simple forms of associative meaning and connotations as well as illocutionary forms of expression, but lacks kinds of more complex forms of combinatorial semantics \citep[see the discussion of][]{koelsch2011a, slevc2011meaning, fitchGingras2011, davies2011, reich2011}. 
However, it is possible to conceive of complex forms of musical tension that involve nested patterns of expectancy and prolongation as an abstract secondary structure, and motivate syntactic structures at least in Western tonal music, and in analogy would require characterising a transducer mapping syntactic structure and corresponding structures of musical tension in future research.

\subsection{Semantics in animal song}

Similarly, there have been debates about the semantic content of animal communication. 
There are a few reported cases of potential compositional semantics in animal communication \citep{vonFrischArnoldZuberbuehler}, but these concern sequences of only two elements and thus do not come close to needing the expressiveness of finite-state or more complex transducers. 
For all animal vocalisations that have non-trivial structure, such as the songs of nightingales \citep{weiss2014use}, blackbirds \citep{todtWolffgram1975, ten2013analyzing}, pied butcherbirds \citep{taylor2011australian} 
or humpback whales \citep{payne1971songs, payne1985large}, it is commonly assumed that no combinatorial semantics underlies it. 
However, it is important to note that the ubiquitous claim that animal songs do not have combinatorial, semantic content is actually based on few to no experimental data.  
As long as the necessary experiments are not designed and performed, the absence of evidence of semantic content should not be taken as evidence of absence. 

If animal songs 
do indeed lack semanticity they would be more analogous to human music than to human language.
The analogy to music would then not primarily be based on the surface similarity to music on the level of the communicative medium (use of pitch, timbre, rhythm or dynamics), but on functional considerations such as that they do not constitute a medium to convey types of (propositional) semantics or simpler forms of meaning, but are instances of comparably free play with form and displays of creativity \citep{wiggins2015evolutionary}.  

\subsection{A music-language continuum?}

Does this view on music-animal song analogies have any relevance for the study of language? 
There are reasons to argue it does, because music and human language may be regarded as constituting a continuum of forms of communication that is distinguished in terms of specificity of meaning \citep{brown2000musilanguage, crossWoodruff2009}.
Consider, for instance, several forms of language that may be considered closer to a `musical use' in terms of their use pitch, rhythm, meter, and semantics, such as motherese, prayers, mantras, poetry, and nursery rhymes, as well as perhaps forms of the utterance ``huh'' \citep[see][]{dingemanseTorreiraEnfield2013}.

Animal vocalisations may be motivated by forms of meaning (that are not necessarily comparable with combinatorial semantics), such as expressing aggression or submission, warning of predators, group cohesion, or social contagion, or they may constitute free play of form for display of creativity, for instance (but not necessarily), in the context of reproduction. 
Given that structure and structure building moving from the language end to the music end is less constrained by semantic forms, more richness of structural play and creativity is expected to occur on the musical side \citep{wiggins2015evolutionary}.

\section{Dealing with gradations: adding continuous-valued variables}

An entirely different approach to modelling natural language - parallel to the symbolic one employed by the Chomsky hierarchy - is 
one where the symbols and categories of the CH are replaced by vectors and the rules are projections in a vector space (implicitly) defined in matrix vector algebra.
Thus, instead of having a rule `X $\rightarrow$ Y Z', where X, Y and Z are symbolic objects (such as a `prepositional phrase' (PP) in linguistics, or a motif in a zebra finch song), we treat X, Y and Z as $n$-dimensional vectors of numbers (which can be binary, integer, rational or real numbers; for example [0, 1, 0\ldots] or [0.453, 0.3333, -0.211,\ldots]) and `$\rightarrow$' becomes an operation on vectors that describes how the vector for $Z$ can be computed given the vectors for $X$ and $Y$.
Vector grammars offer a natural way to model similarity between words and phrases, which can be defined as their distance in the vector space.
Consequently, as one can compute how close a vector is to its prototypical version, vector grammars can straightforwardly deal with noisy data, and exhibit a gradual decrease of performance when inputs become longer or noisier, both properties that are attractive for cognitive models of language.

\subsection{Vector grammars and connectionism}

Vector grammars bear a close relation to connectionist neural network
models of linguistic structure that were introduced in the 1990s \citep{elman1990, pollack1990}.
After being practically abandoned as models of linguistic structure for over a decade, neural networks are experiencing a new wave of excitement in computational linguistics, following some successes with learning such grammars from data for practical natural language processing tasks, such as next word prediction \citep{mikolovetal2010}, sentiment analysis \citep{socheretal2010, socheretal2013, leZuidema2014}, generating paraphrases \citep{le2014distributed, iyyer2014neural} and machine translation \citep{bahdanau2014neural}.
As they can straightforwardly deal with phenomena that are continuous in nature (such as loudness, pitch variation or beat) as well as conveniently handle multiple streams at the same time, vector grammars or neural networks - although not frequently applied in this field - also seem very suitable to model music \citep[see][for some examples of recent work in which neural network models are used to model aspects of music]{cherlahybrid, spiliopoulou2011comparing}. 

Whether neural network models are fundamentally up to the task of modelling structure in language and what they can teach us about the nature of mental processes and representations (if anything at all) has been the topic of a longstanding (heated) debate \citep[some influential papers are][]{fodor1988connectionism, pollack1990, rumelhart1986pdp, pinker1988connections}.
Whichever side one favors in this debate, it seems undoubtedly true that the successes of neural networks in performing natural language processing tasks are difficult to interpret and that the underlying mechanisms are difficult to characterise in terms of the structure building operations familiar from the CH tradition.
Part of this difficulty comes from the fact that neural network models are typically trained on approximating a particular input-output relation (`end-to-end') and do not explicitly model structure. 
Although one might argue that many end-to-end tasks require (implicit) knowledge about the underlying structure of language, it is not obvious what this structural knowledge actually entails.
Analysing the internal dynamics to interpret how solutions are encoded in the vector space is notoriously hard for networks that have more than a couple nodes and the resulting systems - a few exceptions aside \citep[e.g.,][]{karpathy2015visualizing} - often remain black boxes.
Furthermore, finding the right vectors and operations that encode a certain task is a complicated task; the research focus is therefore  typically more on finding optimisation techniques to more effectively search through the tremendous space of possibilities than on interpretation \citep[e.g.,][]{zeiler2012adadelta, kingma2014adam, hochreiter1997long, chung2015gated}. 

\subsection{The expressivity of vector grammars}

\begin{figure}
\centering
\begin{tabular}{cc}
    \specialcell{\includegraphics[scale=1.3]{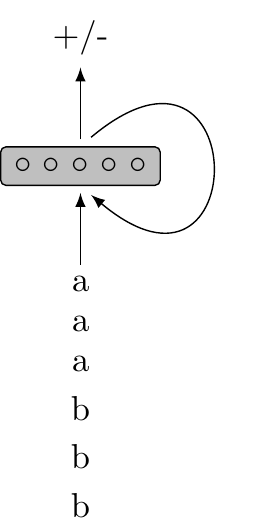}} &
    \specialcell{\includegraphics[scale=1.3]{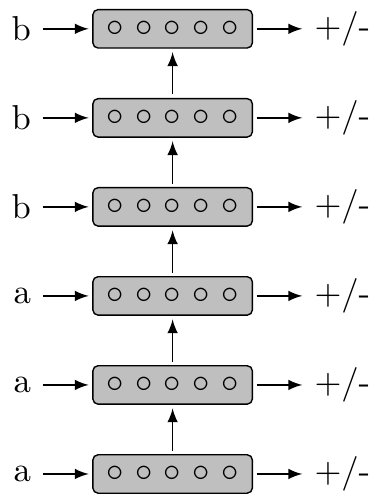}}\\
    (a) & (b)
\end{tabular}

\caption{(a) A simple recurrent network (SRN), equivalent to the one proposed by \cite{elman1990}. The network receives a sequence of inputs (in this case ``a a a b b b'') and outputs whether this is a grammatical sequence (+/-).  The arrows represent so called ``weight matrices'' that define the projections in the vector space used to compute the new vector activations from the previous ones.
NB: traditionally the SRN does not classify whether sequences belong to a certain language, but predicts at every point in the sequence its next element (including the `end of string' marker. Whether a sequence is grammatical can then be evaluated by checking if the network predicted the right symbol at every point in the sequence where this was actually possible (thus, in the case of $a^nb^n$, predicting correctly the end of the string, as well as all the \textit{b}'s but the first one).
(b) The same network, but unfolded over time. On an architectural level, the SRN is similar to an FSA.
}\label{fig:SRN}

\end{figure}

The focus or difficulties in the field aside, however, one can observe that the expressivity of connectionist models reduces the need for more complex architectures (such as MCSG's), as vector grammars are computationally much more expressive than symbolic systems with similar architectures.
For instance, \cite{rodriguez2001} demonstrated that a simple recurrent network (or SRN, on an architectural level similar to an FSA) can implement the counter language $a^nb^n$, a prime example of a context-free language (see Figure \ref{fig:SRN}).
Theoretically, one can prove that an SRN with a non-linear activation function is a Turing complete system that can implement any arbitrary input-output mapping \citep{siegelmann1992computational}.
Although it is not well understood what this means in practice - we lack methods to find the parameters to do so, appropriate techniques to understand potential solutions and arguably even suitable input-output pairs - the theoretical possibility nevertheless calls into question the \textit{a priori} plausibility of the hypothesis that context-freeness is uniquely human.
Rodriguez's results demonstrate a continuum between finite-state and (at least some) context-free languages, which cast doubt on the validity of the focus on architectural constraints on structure building operations that dominates the CH.
As such, while much theoretical work exploring their expressive power is still necessary, vector grammars provide another motivation to move on to probabilistic, non-symbolic models that go beyond the constraints of the CH.

\section{Discussion}\label{sec:discussion}

We have discussed different formal models of syntactic structure building, building blocks and functional motivations of structure in language, music and bird song. 
We aimed to lay a common ground for future formal and empirical research addressing questions about the cognitive mechanisms underlying structure in each of these domains, and about commonalities as well as differences between music and language, and between species.
This chapter can thus be seen as a long-overdue effort to bring theoretical approaches in music and animal vocalisation into common terms that can be compared with approaches established in formal and computational linguistics, 
complementing the literature responding to Hauser, Chomksy and Fitch's provocative hypothesis concerning the `exceptional' role of the human cognitive/communicative abilities.

Our journey through the computational models of structure building - from Shannon’s n-grams, via the CH to vector grammars and models beyond the CH - has uncovered many useful models for how sequences of sound might be generated and processed. 
We arrived at a discussion of recent models that add probabilities, semantics and graded categories to classical formal grammars. Graded category models, which we called vector grammars, 
link formal grammar and neural network approaches and add the power to deal with structures that are inherently continuous.
An important finding using such vector grammars is that one relatively simple architecture can predict sequences of different complexity in the CH and therefore have the potential to undermine assumptions concerning categorically different cognitive capacities between human and animal forms of communication \citep{chomsky1980rules,hauser2002faculty}.

Perhaps the most important lesson we can draw from comparing models for structure building, is that the chosen level of description determines much about the type of conclusions that can be drawn, and that there is no single `true' level of description: the choice of model should therefore depend strongly on the question under investigation.
This is true within a certain paradigm (such as the choice of basic building blocks, or the level of comparison) and between paradigms.
Most comparative research to structure building compares words in language with notes in music and animal song, and sentences to songs.
But this ignores the potential structure in bouts of songs.
If songs were in fact to be compared with \textit{words} rather than sentences, such models would be comparing the bird's phonology with human syntax \citep{yip2006search}.

Moreover, the choice of \textit{model} determines which aspects of structure building we are comparing across domains.
What should be considered in this case is not which model is better in itself, but which model is better for a certain purpose.
For instance, fully symbolic models from the CH provide a useful perspective for the comparison of different theoretical approaches and predictions concerning properties of sets of sequences, but at the same time, they might not necessarily tell us much about the cognitive mechanisms that underlie learning, processing or generating such sets of sequences.
This means that even if we can show that animal song and human language are of a different complexity class in the CH, we cannot automatically assume that there is a qualitative difference in
the cognitive capacities of humans and non-human animals, as demonstrated by the relatively simple neural network architectures that can predict sequences of different complexity in the CH.

\section{Acknowledgements}

This chapter is a thoroughly revised version of Rohrmeier et al. 2015.
We thank the reviewers and editors who offered advice and comments on the different versions of the manuscript.
Special thanks go to the late Remko Scha, whose skepticism and encouragement have indirectly had much influence on our discussion of the cognitive relevance of the Chomsky hierarchy.

\bibliographystyle{plainnat}
\bibliography{library}

\end{document}